\documentclass[letterpaper]{article} 
\usepackage{aaai25}  
\usepackage{times}  
\usepackage{helvet}  
\usepackage{courier}  
\usepackage[hyphens]{url}  
\usepackage{graphicx} 
\usepackage{natbib}  
\usepackage{caption} 
\usepackage{algorithm}
\usepackage{algorithmic}
\usepackage{booktabs}
\usepackage{subcaption}
\usepackage{multirow}
\usepackage{newfloat}
\usepackage{listings}
\DeclareCaptionStyle{ruled}{labelfont=normalfont,labelsep=colon,strut=off} 
\floatstyle{ruled}
\newfloat{listing}{tb}{lst}{}
\floatname{listing}{Listing}
\usepackage{amsmath}

\title{AI Guide Dog: Egocentric Path Prediction on Smartphone}
\author {
    Aishwarya Jadhav\textsuperscript{\rm 1, \rm 3, \rm 4}, \\
    Jeffery Cao\textsuperscript{\rm 1, \rm 3}, 
    Abhishree Shetty\textsuperscript{\rm 1, \rm 3}, 
    Urvashi Priyam Kumar\textsuperscript{\rm 1, \rm 3}, 
    Aditi Sharma\textsuperscript{\rm 1, \rm 3}, \\
    Ben Sukboontip\textsuperscript{\rm 1, \rm 3}, 
    Jayant Sravan Tamarapalli\textsuperscript{\rm 1, \rm 3}, 
    Jingyi Zhang\textsuperscript{\rm 1, \rm 3},
    Anirudh Koul\textsuperscript{\rm 2}
}
\affiliations {
    \textsuperscript{\rm 1}Language Technologies Institute, Carnegie Mellon University, Pittsburgh\\
    \textsuperscript{\rm 2}Pinterest\\
}

\begin{document}

\maketitle

\footnotetext[3]{Authors currently at (in order) Waymo, Celonis, Microsoft, Microsoft, ThinkTrends, Google, Amazon, Snap}
\footnotetext[4]{Corresponding author: anjadhav@alumni.cmu.edu}

\begin{abstract}
This paper presents AI Guide Dog (AIGD), a lightweight egocentric (first-person) navigation system for visually impaired users, designed for real-time deployment on smartphones. AIGD employs a vision-only multi-label classification approach to predict directional commands, ensuring safe navigation across diverse environments. We introduce a novel technique for goal-based outdoor navigation by integrating GPS signals and high-level directions, while also handling uncertain multi-path predictions for destination-free indoor navigation. As the first navigation assistance system to handle both goal-oriented and exploratory navigation across indoor and outdoor settings, AIGD establishes a new benchmark in blind navigation. We present methods, datasets, evaluations, and deployment insights to encourage further innovations in assistive navigation systems. 
\end{abstract}

%

\section{Introduction}
\label{sec:intro}

Navigation assistance systems for visually impaired individuals have been studied for several decades \cite{Dimitrios}, yet their widespread adoption remains limited due to (1) reliance on expensive, custom-built devices, (2) the lack of efficient, robust models that ensure user safety, and (3) limited user trust and convenience.

Existing systems often rely on expensive devices with built-in sensors like LiDAR, or laser scanners \cite{Wang, Wachaja, Hesch}, for 3D mapping, or IMUs, gyroscopes, and pedometers \cite{Lee, Hesch} for localizing user position and orientation. While accurate, these systems are bulky and prohibitively expensive, limiting their accessibility. To address these challenges, we propose a lightweight system leveraging only video feed from a smartphone camera worn near the user’s chest. This video stream is processed in real-time by an on-device model to generate navigation instructions, which are translated into audio cues for the user. This facilitates accessibility and ease of adoption while maintaining robust performance.

\begin{figure}[t]
  \centering
   \includegraphics[height=0.62\linewidth]{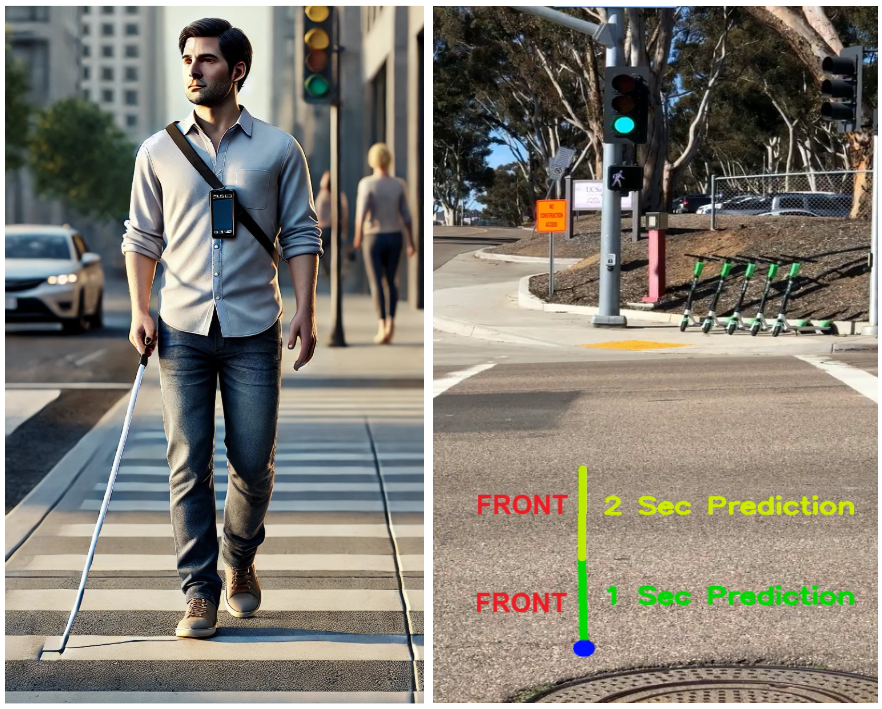}
   \caption{AIGD requires just a smartphone camera and predicts future navigation direction labels.}
   \label{fig:main_fig}
\end{figure}

Most prior work \cite{Diffusion, Multimodal, SPark, Yagi} on blind navigation formulates it as a trajectory forecasting problem, predicting precise 3D locations and orientations of the user. However, this approach typically reflects the behavior of sighted individuals, who navigate by dynamically avoiding obstacles and other pedestrians. Blind navigation is fundamentally different: visually impaired individuals typically prefer retaining canes \cite{Ohn-Bar}, even when assisted by robotic systems. Canes help detect obstacles and signal others to yield space, facilitating smoother navigation.Thus, the user-environment dynamics of blind individuals differ significantly from those of sighted users. 

This insight allows us to simplify the navigation task into an egocentric path prediction problem, where the system predicts all possible future user navigation actions—such as continuing straight, turning left, or turning right. This abstraction avoids the uncertainties of precise trajectory prediction and instead focuses on the user’s heading direction and actions relative to their environment. We adopt a multi-label classification approach to accommodate multiple possible navigation directions and enable easy translation of the model’s outputs to clear and actionable commands for users. While limited prior work \cite{Wang, Krishnacam} explored similar ideas, they were restricted to single-class predictions in constrained environments.

Despite advancements, existing blind navigation models lack the robustness and accuracy needed for reliable real-world use. To build trust, the system must support blind users across diverse scenarios—avoiding collisions in cluttered indoor spaces, as well as facilitating outdoor navigation with goal-based guidance from GPS-enabled apps like Google Maps. However, instructions from such apps (e.g., ``Turn left at W. 4th St.'' \cite{gdirection}) are often impractical for blind users in dynamic outdoor environments. Consequently, most prior work focuses on no-goal \cite{Diffusion, Multimodal} or fixed-path \cite{Ohn-Bar} navigation, typically limited to either indoor or outdoor settings. AIGD bridges this gap by enabling both exploratory and goal-based navigation, allowing users to navigate freely or follow specific destinations. AIGD is the first system to handle scenarios across indoor and outdoor environments, with and without the \textit{intent} of reaching a final destination, while also accounting for multiple possible directions in the absence of a goal.

First-person camera inputs face challenges like jitter, blur, limited fields of view, and variations in smartphone quality, camera positioning, and walking speeds. Although the slower, deliberate movements typical of visually impaired users (average walking speed: 0.72 \text{m/s} \cite{Zhang}) reduce extreme ego-motion effects, these challenges persist. Furthermore, real-world navigation data exhibits significant imbalance, with far more forward movement instances than turning actions. These observations inform our data collection, curation, modeling and deployment processes.
\newline
\newline
\noindent Our key contributions include:
\begin{enumerate}
    \item \textbf{A robust, lightweight multi-label classification model} addressing turn class imbalance, and effective across scenarios with or without destination intent.
 \item \textbf{A methodology} for integrating destination and high-level direction signals into vision-only prediction models, validated by extensive experiments.
    \item \textbf{An open-source dataset} comprising egocentric videos and associated mobile sensor data collected across diverse scenes and participants, facilitating future research.
    \item \textbf{A low-latency smartphone app} deploying the model for real-world navigation assistance.
\end{enumerate}

\section{Related Work}
\label{sec:relatedwork}
\subsection*{Blind Navigation Assistance Systems}
Previous blind assistance systems have primarily utilized either wearable devices or robotic helpers. Wearable systems incorporate body-mounted sensors \cite{abidi} (e.g., feet, knees, or waist) and rely on standalone devices, like laptops in backpacks \cite{Lee2}, smartphones \cite{Sato, Pawar} or tablets \cite{Isana} for computations. For instance, \citeauthor{Lee2} \shortcite{Lee2} developed a head-mounted RGB-D camera system for egomotion estimation and obstacle-aware path planning, providing tactile feedback through a haptic vest. Similarly, \citeauthor{Wang} \shortcite{Wang} introduced a wearable system with a structured light camera generating point clouds and conveying navigation instructions via vibration motors and Braille. 

Robotic helpers, such as smart canes \cite{Hesch, Gupta, Yang} or suitcase-mounted devices \cite{Manglik}, act as robotic guide dogs. For instance, \citeauthor{Wachaja} \shortcite{Wachaja} proposed a robotic walker with laser rangefinders and servo motors for egomotion estimation and obstacle detection, while ISANA \cite{Isana} employs a Google Tango tablet for computation and an onboard RGB-D camera for obstacle avoidance, providing multimodal feedback through speech, audio, and haptics.

\subsection*{Egocentric Navigation}
Egocentric navigation comprises trajectory forecasting, which predicts future 2D/3D positions, and path prediction, which identifies discrete directional actions (e.g., left, right, forward), with related research extending into goal-based navigation. Trajectory forecasting, while extensively studied for vehicles, has seen limited attention for human navigation. \citeauthor{Yagi} \shortcite{Yagi} proposed a convolution-deconvolution network using pedestrian poses and ego-motion, while others have used GRU-CNN \cite{Styles} and LSTM encoder-decoder models \cite{Qiu} to predict human trajectories in indoor environments. More recent methods leverage multimodal transformers \cite{Multimodal} and diffusion models \cite{Diffusion} to incorporate scene semantics and past trajectories for future position predictions.

Egocentric path prediction methods include \citeauthor{Lee} \shortcite{Lee}, which generates 3D occupancy maps and utilizes D*-Lite planning for four directional cues (straight, left, right, stop); \citeauthor{Krishnacam} \shortcite{Krishnacam}, which uses a fine-tuned CNN to predict discrete motion classes from single camera images; and \citeauthor{Ohn-Bar} \shortcite{Ohn-Bar}, which developed a dynamic path planning policy personalized to user reaction times, providing discrete localized guidance based on global pre-planned layouts.

Most goal-based navigation research focuses on robotics \cite{robotsurvey} and autonomous vehicles \cite{Aradi}, relying on dynamic path planning and reinforcement learning. However, these approaches are unsuitable for modeling human behavior, particularly for blind users, who face unique social and reaction constraints.

To the best of our knowledge, AIGD is the first system to generalize navigation for blind users across diverse scenarios. Our approach is motivated by the unique requirements of this use-case, allowing us to scope the problem to a limited set of instruction classes while incorporating goal-based navigation and directional uncertainty, without relying on complex dynamic planning or reinforcement learning. This ensures a solution that is both practical and user-centric.

\section{Method}
\label{sec:method}

\subsection{System Overview}

\begin{figure}
  \centering
   \includegraphics[width=\linewidth]{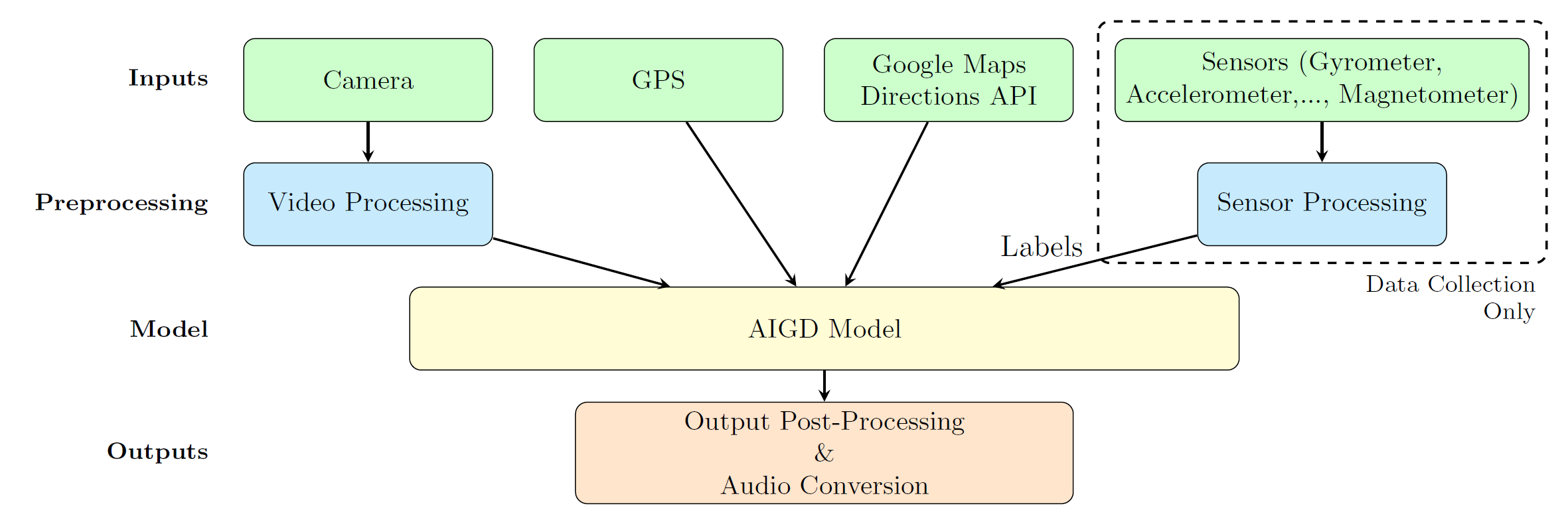}
   \caption{System Architecture for AIGD.}
   \label{fig:sys_arch}
\end{figure}

Our system features a smartphone app running a lightweight, real-time model on-device, using video input from the device’s camera, and optionally GPS and Google Maps data for destination-based navigation. Sensor data (e.g., accelerometer, gyrometer) is used only during data collection for prediction label generation, not for model inference in the deployed user app. Navigation predictions are post-processed into audio instructions for the user. Fig. \ref{fig:sys_arch} depicts the overall system architecture.

\subsection{Problem Definition}

We model this task as a multi-label classification problem, where, for each frame sampled from the smartphone camera stream, the system predicts the user’s heading direction one second into the future. Specifically, based on the current scene and, optionally, past frames, the model outputs classification scores for three possible directions (FRONT, LEFT, RIGHT) the user could take in the next second. The one-second future horizon is informed by studies on walking speeds \cite{Zhang} and average reaction times \cite{Bhirud} specific to our blind user base. 

\begin{figure}
  \centering
   \includegraphics[width=\linewidth]{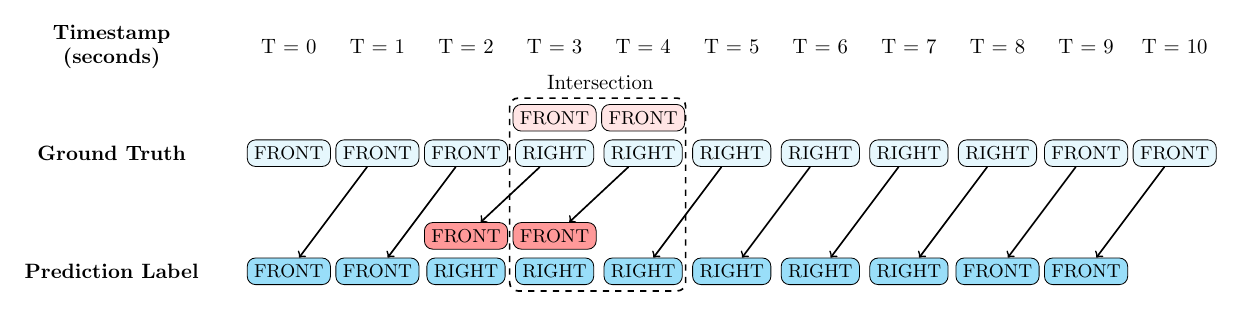}
   \caption{Labeling Scheme for frames sampled at 1 FPS. Red blocks denote other walkable directions at intersection.}
   \label{fig:labels}
\end{figure}

Multiple turn labels are generated only in scenarios without destination intent, typically at intersections or when pathways diverge. In such cases, the model must predict all possible directions one second before the turn begins. During the turn, it must predict the active turn direction (LEFT/RIGHT), and finally transition to predicting FRONT, with high confidence, as the turn concludes. Fig. \ref{fig:labels} illustrates this with frames sampled at 1 FPS. For free-space navigation without obstacles, our labeling scheme conditions the model to output only the FRONT direction.

\subsection{Dataset}
At the time of our study, existing egocentric walking datasets lacked the diversity of scenarios and the sensor or GPS data needed for our use case. Besides, most were confined to specific environments (e.g., labs, offices), and relied on specialized cameras or hardware. 

To better reflect the real-world conditions of our app’s usage, we collected a custom dataset using smartphone cameras, aiming to capture the walking speeds and styles of visually impaired individuals. To constrain the study, we used iPhone 13 with the AIGD data collection app installed. Eight participants, primarily graduate students and tech interns, collected data in semi-crowded spaces with stationary obstacles and people. To simulate social navigation interactions similar to those experienced by our target users, participants wore black glasses and carried the smartphone on a lanyard near their chest, walking slowly and cautiously. This setup also captured variations in first-person camera movements, camera positionings, and fields of view. 

Data was collected for diverse indoor and outdoor scenes across Pittsburgh, Seattle, and the Bay Area, as described in Tab. \ref{tab:data_split}, focussing on everyday venues outside users’ homes or offices. Indoor environments included well-lit spaces with numerous aisles, such as grocery stores, to increase the frequency of left and right turns. Since there is no destination for these, participants were instructed to turn at every available opportunity. Outdoor data was collected in parks with winding pathways and city streets during daytime. All walking paths were unscripted and unplanned, with each video capturing a single scene and lasting 2 to 10 minutes.

\begin{figure}[t]
  \centering
   \includegraphics[width=0.95\linewidth]{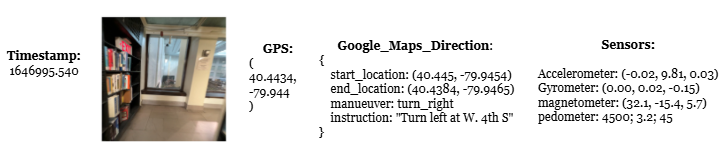}
   \caption{Data example for one timestep.}
   \label{fig:data_example}
\end{figure}

In total, the dataset includes 57 hours of walking data captured at 30 fps. It includes video from smartphone cameras and sensor data (accelerometer, gyrometer, magnetometer, pedometer) captured at 0.1-second intervals. For outdoor scenes, GPS locations and directional data from the Google Maps API were also logged. Fig. \ref{fig:data_example} shows a raw example record.

Sensor data was processed through denoising, reference transformations, and windowing to generate ground truth labels. It was then timestamp-aligned with video frames downsampled to 2 fps and converted to 128×128 grayscale. GPS and high-level destination directions were aligned with these records, where available. Each data example consists of a frame, its label, the preceding 5 seconds (10 frames) as context, and associated GPS and direction features for all frames. The past context frames help inform future predictions by implicitly capturing the user's walking speed and reaction time. The dataset comprises 392,580 examples, divided into training, validation, and test sets in a 60:20:20 ratio, ensuring no overlap of scenes across splits, as summarized in  Table \ref{tab:data_split}. 

We open-source a subset of the collected egocentric videos and associated mobile sensor data, where release consent has been obtained from relevant authorities, ensuring compliance with ethical and privacy regulations. Our dataset is available online.\footnote{\url{http://bit.ly/41h7jJn}}

\begin{table}[t]
\centering
\small
\begin{tabular}{@{}lllc@{}}
\toprule
 & & \textbf{Scene} & \textbf{Data Split} \\ 
\midrule
\multirow{4}{*}{\textbf{Indoor}}
    &             & Library 1 & Train \\
    & 32 hours;    & CMU Hall & Train \\
    & 220k     & Library 2 & Validation \\
    & examples                    & Grocery Store & Test \\ 
\midrule
\multirow{5}{*}{\textbf{Outdoor}}
    &             & Pittsburgh Street 1 & Train \\
    & 25 hours;       & Park (70\% videos) & Train \\
    &  172k   & Pittsburgh Street 2 & Validation \\
    &  examples  & Seattle Street 1 & Test \\
    &                     & Park (30\% videos) & Test \\
\bottomrule
\end{tabular}
\caption{Data Splits}
\label{tab:data_split}
\end{table}

\subsubsection{Ground Truth Labels}

Labels for each sampled frame were derived from sensor data. Various methods, including accelerometer, compass, GPS, and optical flow-based approaches were evaluated for turn label calculation. Among these, the compass-based method consistently yielded the most accurate results, particularly at slower walking speeds, proving less noisy than accelerometers and more precise than GPS for heading estimation. Details of these approaches, evaluations and parameter tuning, follow \cite{aigd0}. Below is a brief overview.

For each data point, the turning angle is computed by comparing the agent's facing direction over a 0.5-second interval. Angles above $5^\circ$ indicate a RIGHT turn, below $-5^\circ$ a LEFT turn, and within $\pm5^\circ$ represent FRONT movement. The $5^\circ$ threshold, an adjustable hyperparameter, controls sensitivity to minor orientation changes. 

For indoor scenarios without destination intent, auto-calculated turn labels were manually re-labeled to ensure all possible turn directions at  intersections were captured. Multiple labels were assigned to the initial 2 seconds of each turn.

\subsection{Models}
\label{sec:models}

This section outlines the models used to validate our proposed problem formulation and intent integration methodology. For no-destination (no-intent) navigation, we implement multi-label classification models, including simple baselines such as CNN and ConvLSTM, commonly employed in prior work \cite{Styles, Qiu, Krishnacam}, as well as the advanced PredRNN model \cite{Predrnn}. We then extend these no-intent models by incorporating destination intent features and GPS signals to enable goal-conditioned predictions.

\textbf{CNN: }
Following \citeauthor{Krishnacam} \shortcite{Krishnacam}, we finetuned a ResNet34 model with a linear classification head to encode individual frames. This baseline model only considers per-frame information, disregarding the temporal context provided by preceding frames. 

\textbf{ConvLSTM: }
The ConvLSTM \cite{Convlstm} architecture, described in Fig. \ref{fig:conv_lstm}, designed for spatiotemporal prediction, serves as our temporal baseline. It replaces the input-to-state and state-to-state transitions of standard LSTMs with convolutional structures, which allow it to effectively leverage the visual information in the preceding context frames for improved predictions. However, it is computationally intensive and susceptible to overfitting, particularly with limited fine-tuning data. 

\textbf{PredRNN:}
PredRNN \cite{Predrnn} utilizes spatiotemporal LSTM units to model sequential dependencies in video data, and is widely used for future frame prediction tasks \cite{Jadhav, Ma}. We explore PredRNN’s ability to model complex short-term dynamics for our future direction prediction use-case. However, like ConvLSTM, PredRNN is computationally demanding for both training and inference, with even higher latency due to its increased architectural complexity.

\subsection*{Intent-based Navigation}
For outdoor navigation, directions from Google Maps provide high-level guidance by localizing the user via GPS. However, GPS accuracy ($\sim$ 4.9 \text{meters} \cite{gps}) is insufficient for precise local navigation. To address this, the model must predict local directions, and the corresponding actions to take in the next second, that are aligned with the high-level Maps directions and the user’s GPS history.

In this work, we use the \textit{Google Maps Directions API} \cite{gdirection}, which provides step-by-step walking instructions for specified start and destination addresses. For the ``walking'' mode, the API returns an array of steps, each containing a \texttt{start\_location} (latitude-longitude), \texttt{end\_location}, and a \texttt{maneuver} or action to take at the \texttt{end\_location}. Each step corresponds to a travel segment. At each timestep, we pick the appropriate segment to retrieve the manueuver from based on the user's GPS position and the segment's \texttt{start\_location} and \texttt{end\_location}. Relevant maneuvers for walking include \texttt{turn-slight-left}, \texttt{turn-sharp-left}, \texttt{turn-left}, \texttt{turn-slight-right}, \texttt{turn-sharp-right}, \texttt{turn-right}, and \texttt{straight}. We one-hot-encode the maneuver values and append them with the latitude and longitude from the \texttt{start\_location} and \texttt{end\_location} fields to create an \textbf{\textit{Intent Feature}} for each step. The Intent Feature is then combined with the user’s current GPS coordinates and passed through a linear embedding layer to generate the \textbf{\textit{Intent Embedding}} vector, which serves as a high-level intent conditioning input for the model.

The following sections detail the modifications made to the baseline CNN and ConvLSTM architectures to incorporate Intent Embeddings. As discussed in the results section \ref{sec:nointentperf}, the performance gains offered by PredRNN are outweighed by its high computational requirements and latency, which are critical factors for smartphone deployment. Hence, it is excluded from the intent-based experiments. Instead, we implement a more efficient, hybrid CNN+LSTM architecture to capture both temporal and intent signals.

\textbf{CNN with Intent}: Intent embeddings are concatenated with CNN-extracted frame embeddings, which are then passed through a 2-layer MLP for prediction.

\begin{figure}[t]
    \centering
    \begin{subfigure}[b]{0.8\linewidth}
        \includegraphics[width=\linewidth]{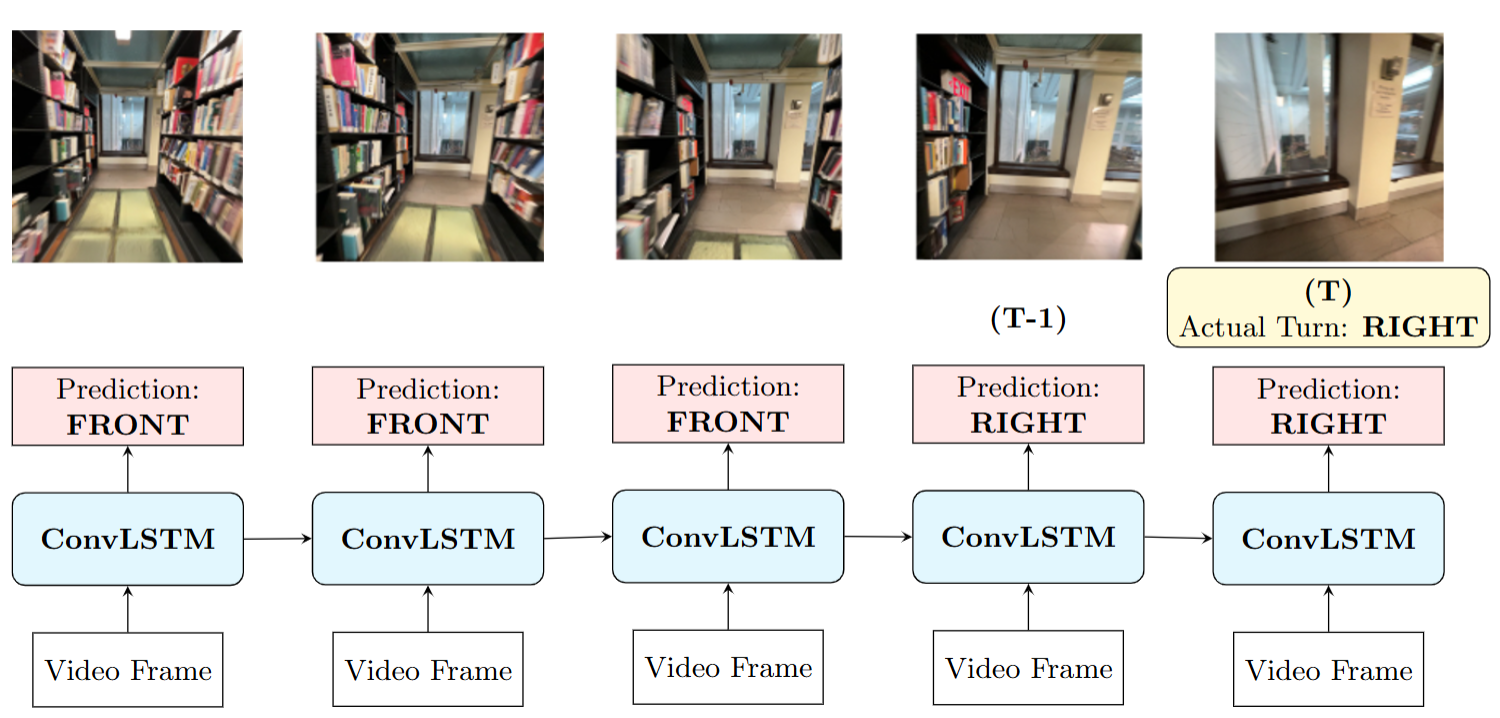}
    \caption{ConvLSTM Baseline}
        \label{fig:conv_lstm}
    \end{subfigure}

    \begin{subfigure}[b]{0.85\linewidth}
        \includegraphics[width=\linewidth]{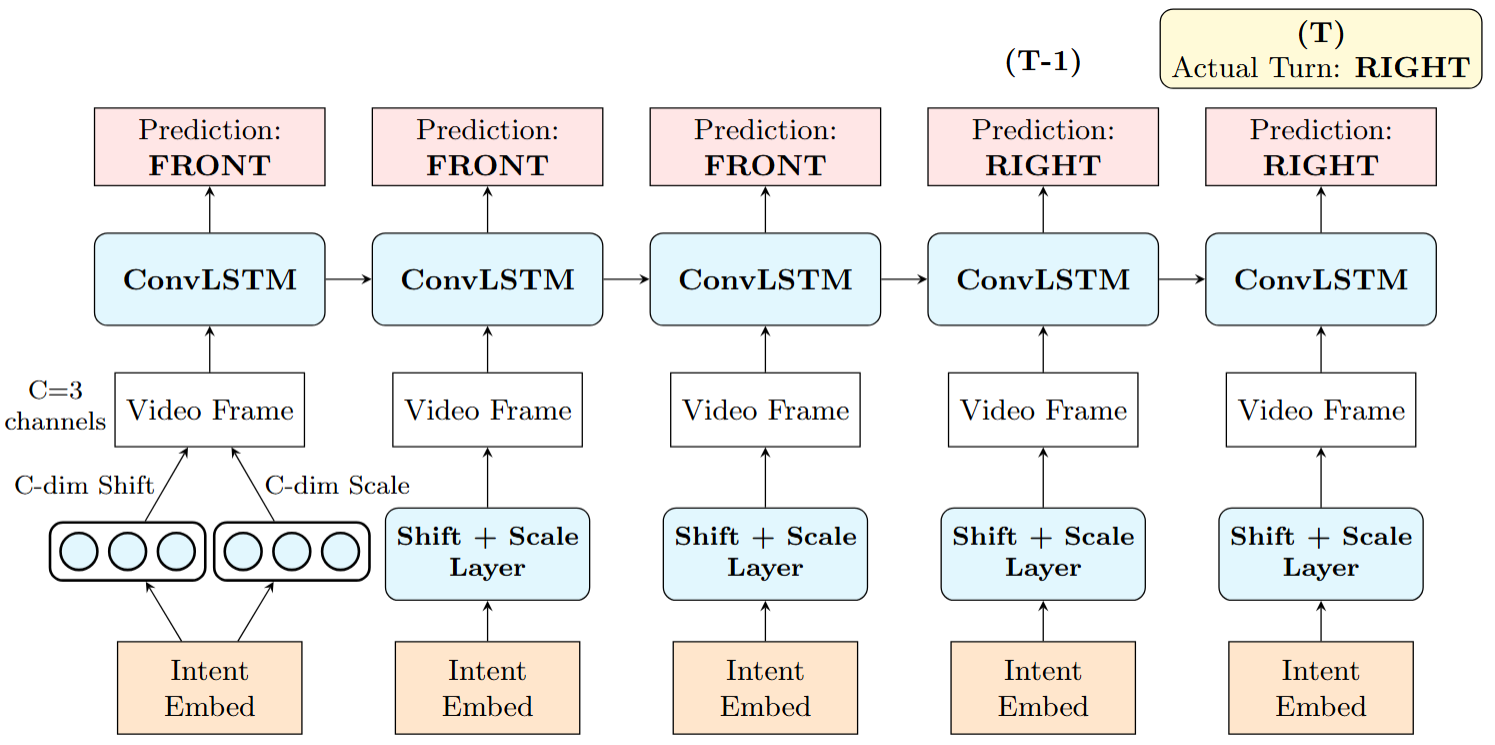}
        \caption{Intent-modified ConvLSTM Model}
        \label{fig:conv_lstm_intent}
    \end{subfigure}
       \caption{ConvLSTM with and without intent modifications}
    \label{fig:conv_lstm_all}
    
\end{figure}

\textbf{ConvLSTM with Intent} (Fig. \ref{fig:conv_lstm_intent}): Intent embeddings are projected down to two C-dimensional vectors, where C=3 corresponds to the number of video frame channels. These vectors are added as shift and scale factors to the frame input channels before passing through the ConvLSTM layers. While we explored other early fusion strategies, this method demonstrated the best performance vs complexity tradeoff.

\begin{figure}[t]
  \centering
   \includegraphics[width=0.9\linewidth]{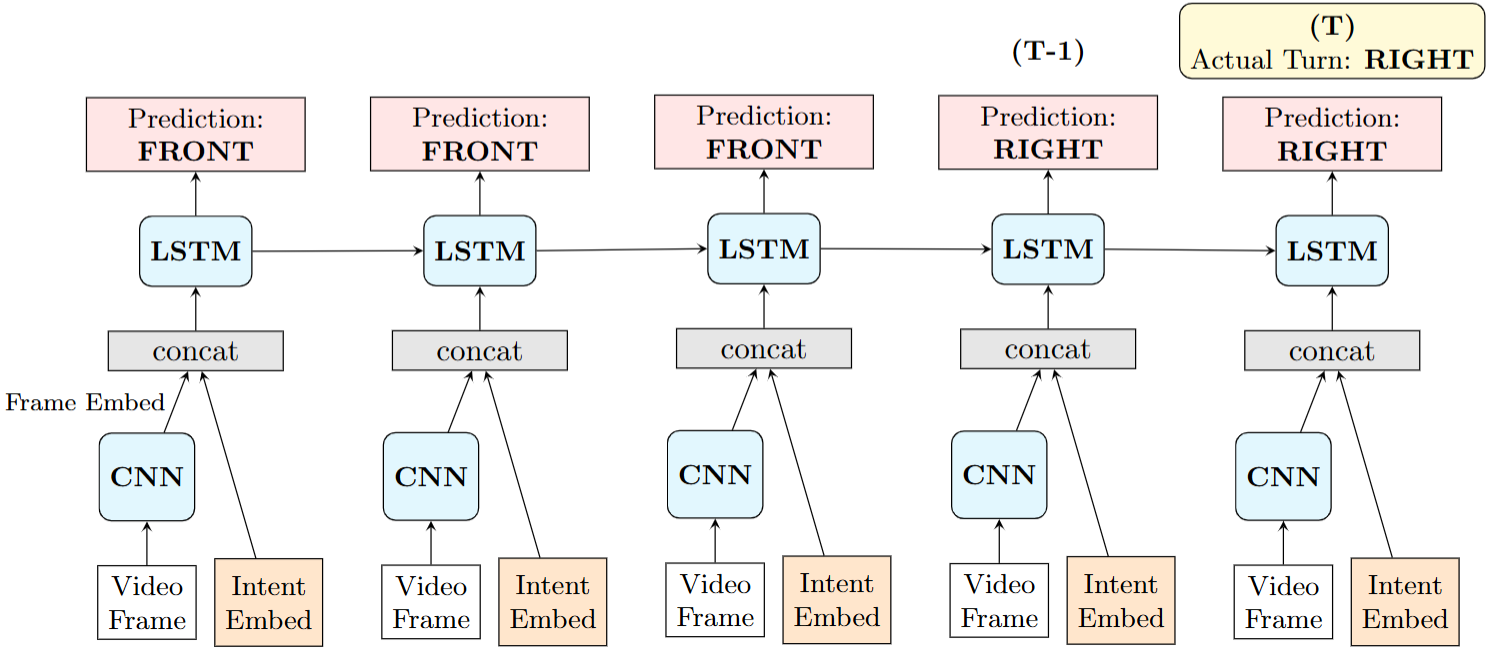}
   \caption{CNN+LSTM with Intent}
   \label{fig:cnn_lsmt_intent}
\end{figure}

\textbf{CNN + LSTM with Intent} (Fig. \ref{fig:cnn_lsmt_intent}): This architecture enhances the CNN+Intent model by replacing the MLP in the final classifier with a 2-layer LSTM. It combines past frame embeddings, extracted via the powerful ResNet feature extractor, with the corresponding timestep's intent embeddings, and uses LSTMs to model past context temporal relationships. Compared to ConvLSTM, CNN+LSTM is more computationally efficient as the LSTM processes lower-dimensional embeddings instead of full image data.

\section{Experiments}
\label{sec:experiments}

\subsection{Label Imbalance}
\begin{figure}[t]
  \centering
   \includegraphics[width=0.6\linewidth, height=0.35\linewidth]{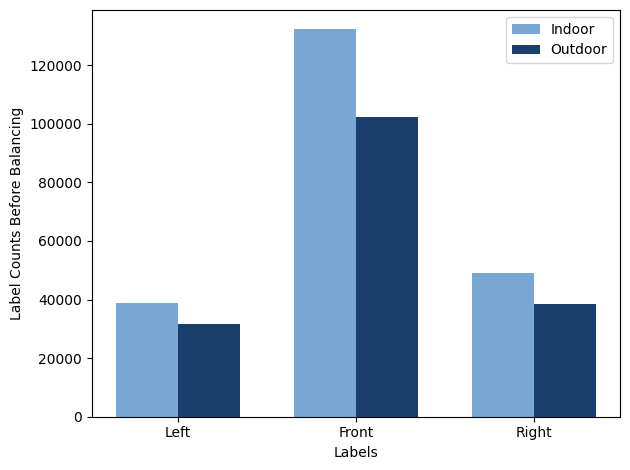}
   \caption{Distribution of class labels in the dataset.}
   \label{fig:imbalance}
\end{figure}

Despite efforts to collect more turn-based data, the dataset remains significantly skewed toward the FRONT label, as seen in Fig. \ref{fig:imbalance}. However, predicting turns (LEFT and RIGHT) is more critical for navigation. To address this imbalance and improve turn prediction, we implemented the following during training:
\begin{enumerate}
    \item \textbf{Minority Oversampling:} Doubled LEFT/RIGHT class examples.
    \item \textbf{Data Augmentation:} Applied transforms described in Tab. \ref{tab:augmentations} with a 20\% probability.
    \item \textbf{Loss Function:}
\end{enumerate}

\begin{table}
  \centering
  \small
  \begin{tabular}{@{}p{2cm} p{2.5cm} p{2.75cm}@{}}
    \toprule
    \textbf{Augmentation} & \textbf{Specs} & \textbf{Purpose} \\
    \midrule
    Translation & 5–25 pixels vertical and horizontal & Camera at varying heights\\
    Color Jitter & 0-20\% HSV & Varying lighting \\
    Random Crop & 5-20 pixel squares & Real-world occlusions \\
    Rotation & -20 to +20 degrees & Camera rotations during walking \\
    Noise & Gaussian or salt-pepper  & Differing camera qualities \\
    Super-pixel & 8-pixel clusters & Bad resolution/noise \\
    \bottomrule
  \end{tabular}
  \caption{Augmentations Settings}
  \label{tab:augmentations}
\end{table}

Oversampling reduces class imbalance but does not fully address the more challenging, yet rarer, turn prediction cases near the start and end of a turn. To mitigate this, we employed Focal Loss \cite{Focal}, which emphasizes harder-to-classify samples by dynamically scaling their loss contribution. We also experimented with Weighted BCE Loss using class weights of 2:2:1 (LEFT:RIGHT:FRONT), which provided minor performance gains. These weights were integrated into our focal loss formulation.

The above sampling, augmentation, and loss function settings were determined through experiments on CNN and ConvLSTM baselines.

\subsection{Experimental Setup}

All models were fine-tuned using the label balancing settings described above. Best available public checkpoints were used to initialize the CNN and ConvLSTM components for both no-intent and intent-based models, as well as PredRNN. The final MLP in the CNN-based models, LSTM in the CNN+LSTM+Intent model and the Intent Embedding layers were trained from scratch. Training was conducted for 30 epochs with a batch size of 64, using the Adam optimizer with a weight decay of 1e-3. The CosineAnnealingLR scheduler was used, with learning rates of 1e-4 for layers trained from scratch and 1e-5 for fine-tuned layers.

\label{sec:ablation}
We conducted ablation studies to assess the impact of different training data configurations. No-intent models are well-suited for indoor scenarios, where GPS and high-level directions are unavailable, but multi-label supervision is provided. In contrast, intent-based models effectively leverage GPS and directional features in outdoor datasets. Hence, we trained the no-intent models exclusively on the indoor dataset and the intent-based models on the outdoor dataset. Performance was evaluated on corresponding test sets and benchmarked against our generalized intent models trained on combined indoor and outdoor datasets.

\subsection{Evaluation Metrics}
We evaluate the models using accuracy, AUC, Precision, Recall, and F1 score. 

\section{Results}

Tab. \ref{tab:fullres} details the performance of models trained on combined indoor and outdoor datasets, evaluated on a test set containing both indoor and outdoor video segments. Tab. \ref{tab:iores} breaks down the AUC evaluations of these models for indoor and outdoor test videos separately. Tab. \ref{tab:ablres} summarizes the results of the training dataset ablations.

For all experiments, the performance of the FRONT label remains largely unchanged by the modifications. Since our use case emphasizes turn (LEFT/RIGHT) classes, the following sections focus only on their performance.

\begin{table*}[t]
\centering
\small
\begin{tabular}{@{}p{3cm}cccccccccccc@{}}
\toprule
\textbf{Model}                  & \multicolumn{4}{c}{\textbf{LEFT}}               & \multicolumn{4}{c}{\textbf{RIGHT}}              & \multicolumn{4}{c}{\textbf{FRONT}}             \\ \cmidrule(lr){2-5} \cmidrule(lr){6-9} \cmidrule(lr){10-13}
                                & AUC & Prec. & Rec. & F1              & AUC & Prec. & Rec. & F1              & AUC & Prec. & Rec. & F1              \\ 
\midrule

CNN                     & 0.571 & 0.610 & 0.543 & 0.5746 & 0.608 & 0.702 & 0.567 & 0.6273 & 0.900 & 0.903 & 0.803 & 0.8501 \\
ConvLSTM                & 0.622 & 0.689 & 0.544 & 0.608  & 0.645 & 0.725 & 0.572 & 0.6395 & 0.908 & 0.900 & 0.810 & 0.8526 \\
PredRNN                 & \textbf{0.636} & 0.708 & 0.549 & \textbf{0.6184} & \textbf{0.657} & 0.752 & 0.570 & \textbf{0.6485} & 0.912 & 0.910 & 0.840 & 0.8736 \\ 
\midrule
CNN + Intent         & 0.588 & 0.619 & 0.548 & 0.5813 & 0.622 & 0.711 & 0.571 & 0.6334 & 0.911 & 0.910 & 0.833 & 0.8698 \\
ConvLSTM + Intent    & 0.638 & 0.706 & 0.556 & 0.6221 & 0.659 & 0.742 & 0.571 & 0.6454 & 0.912 & 0.918 & 0.830 & 0.8718 \\
CNN + LSTM + Intent  & \textbf{0.664} & 0.728 & 0.559 & \textbf{0.6324} & \textbf{0.700} & 0.766 & 0.583 & \textbf{0.6621} & 0.920 & 0.920 & 0.846 & 0.8814 \\ 
\bottomrule

\end{tabular}
\caption{Performance for models trained on combined Indoor + Outdoor training datasets. AUC PR, Precision, Recall and F1 scores are reported on entire the test set (Indoor + Outdoor).}
\label{tab:fullres}
\end{table*}

\begin{table*}
\centering
\small
\begin{tabular}{lcccccc}
\toprule
\textbf{Model} & \multicolumn{3}{c}{\textbf{Indoor}} & \multicolumn{3}{c}{\textbf{Outdoor}} \\ \cmidrule(lr){2-4} \cmidrule(lr){5-7}
               & \textbf{LEFT} & \textbf{RIGHT} & \textbf{FRONT} & \textbf{LEFT} & \textbf{RIGHT} & \textbf{FRONT} \\ 
\midrule
CNN                    & 0.579 & 0.614 & 0.905 & 0.550 & 0.592 & 0.900 \\
ConvLSTM               & 0.628 & 0.649 & 0.909 & 0.609 & 0.634 & 0.905 \\
PredRNN                & 0.641 & 0.662 & 0.918 & 0.621 & 0.645 & 0.910 \\ 
\midrule
CNN with Intent        & 0.577 & 0.613 & 0.900 & 0.607 & 0.638 & 0.914 \\
ConvLSTM with Intent   & 0.632 & 0.649 & 0.911 & 0.651 & 0.672 & 0.914 \\
CNN + LSTM with Intent & \textbf{0.660} & \textbf{0.695} & 0.917 & \textbf{0.671} & \textbf{0.707} & 0.920 \\ 
\bottomrule
\end{tabular}
\caption{Performance of models trained on combined Indoor + Outdoor training datasets, with AUCs reported separately for Indoor and Outdoor test data.}
\label{tab:iores}
\end{table*}

\begin{table}
\centering
\small
\begin{tabular}{@{}p{2cm}p{1.4cm}ccc@{}}
\toprule
\textbf{Model} & \multicolumn{1}{p{1.4cm}}{\textbf{Train} \newline \textbf{/ Test}} & \textbf{LEFT} & \textbf{RIGHT} & \textbf{FRONT} \\ 
\midrule
CNN                    & Indoor  & 0.590 & 0.626 & 0.913 \\
ConvLSTM               & Indoor  & 0.643 & 0.655 & 0.918 \\
PredRNN                 & Indoor  & \textbf{0.667} & \textbf{0.687} & 0.925 \\ 
\midrule
CNN + Intent        & Outdoor & 0.600 & 0.629 & 0.906 \\
\multicolumn{1}{@{}p{2cm}}{ConvLSTM\newline+ Intent}   & Outdoor & 0.640 & 0.659 & 0.908 \\
\multicolumn{1}{@{}p{2cm}}{CNN + LSTM\newline + Intent} & Outdoor & 0.66 & 0.682 & 0.912 \\
\bottomrule
\end{tabular}
\caption{Ablations with different train and test data mixes.}
\label{tab:ablres}
\end{table}

\subsection*{Performance of No-Intent Models}
\label{sec:nointentperf}
Without intent features, ConvLSTM and PredRNN outperform CNN by leveraging temporal context, which is particularly beneficial for path disambiguation in the absence of high-level intent cues. Temporal modeling enhances scene understanding, especially during turns, where past frames provide context about the ongoing action (e.g., turning), and the current frame helps decide whether to continue or conclude the turn.

Among no-intent models, PredRNN achieves the best performance on the benchmark test dataset, surpassing ConvLSTM due to its advanced future frame prediction architecture. Despite its complexity, PredRNN generalizes better than ConvLSTM, exhibiting less overfitting. However, its computational demands outweigh its performance gains, making it an unsuitable candidate for on-device deployment.

No-intent models perform better on indoor video segments than outdoor scenes. This is expected, as indoor datasets provide multi-label supervision for ambiguous scenarios like intersections, while outdoor datasets use single-label annotations for each turn. Consequently, we observe many false positives at outdoor intersections, because the no-intent models cannot leverage the disambiguation provided by the Intent Features.

\subsection*{Effects of Adding Intent Features}
Incorporating intent features and GPS signals enhances the performance of CNN and ConvLSTM models over their no-intent counterparts, as shown in Tab. \ref{tab:fullres}. The gains are particularly significant for outdoor test videos (Tab. \ref{tab:iores}), where intent and GPS signals provide explicit directional cues and help disambiguate turns.

For the indoor test set, the performance difference be-
tween no-intent and intent models is negligible. A slight drop in performance is observed in CNN+Intent model compared to its no-intent version, likely noise. Nonetheless, given the substantial gains observed for outdoor scenarios, the intent models depict a net positive improvement while supporting
both scenarios.

Finally, the CNN+LSTM+Intent model outperforms ConvLSTM+Intent in both evaluation metrics and computational efficiency. This mid-fusion approach surpasses the early fusion strategy in ConvLSTM+Intent by independently extracting frame features while jointly modeling temporal information from frame and intent embeddings. Notably, CNN+LSTM+Intent achieves greater gains on indoor videos compared to CNN+Intent and ConvLSTM+Intent models, reducing the performance gap between indoor and outdoor datasets. This is likely due to the later fusion of modalities in the classification head, which better isolates the contributions of video, and Intent/GPS features.

\subsection*{Training Data Ablations}

In tab. \ref{tab:ablres}, the no-intent models trained exclusively on indoor data outperform those trained on a mix of indoor and outdoor data when evaluated on the indoor test videos. In the absence of ambiguous turn supervision in the outdoor training set, these models learn from the much cleaner indoor training set, effectively capturing the aisle and turn patterns. 

In contrast, intent models trained solely on outdoor data perform worse than those trained on a mix of indoor and outdoor datasets, overfitting the smaller outdoor training set. While indoor test metrics are not significantly enhanced by the intent modifications, including indoor videos in the training data benefits the overall performance of intent models.

\subsection*{Qualitative Analysis}
\begin{figure}[t]
  \centering
   \includegraphics[width=0.7\linewidth]{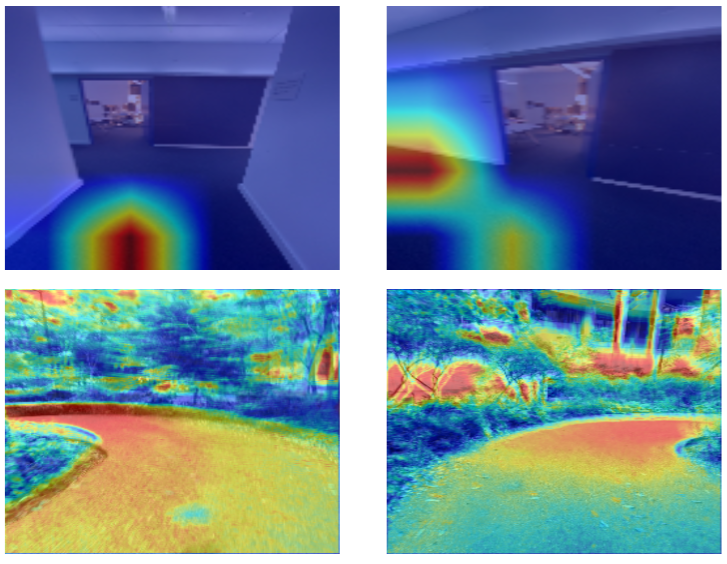}
   \caption{Grad-CAM heatmap visualizations. Top row shows an indoor scene with FRONT and LEFT predictions. Bottom row shows an outdoor scene with turns.}
   \label{fig:gradcam_visualization}
\end{figure}

Fig. \ref{fig:gradcam_visualization} presents GradCAM \cite{Gradcam} visualizations from the CNN model. Even the simple CNN baseline effectively learns path features and curves, enabling it to detect turns in the near future.

\section{Deployment}
\begin{figure}[t]
  \centering
   \includegraphics[width=\linewidth]{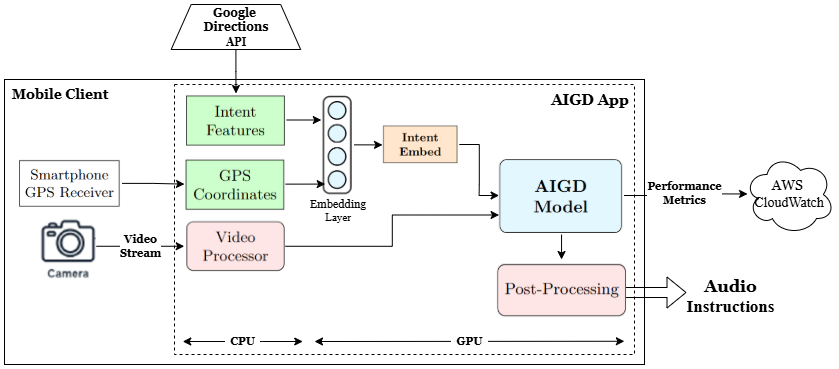}
   \caption{AIGD Deployment Architecture}
   \label{fig:deploy}
\end{figure}

\begin{figure}[t]
  \centering
   \includegraphics[width=0.8\linewidth]{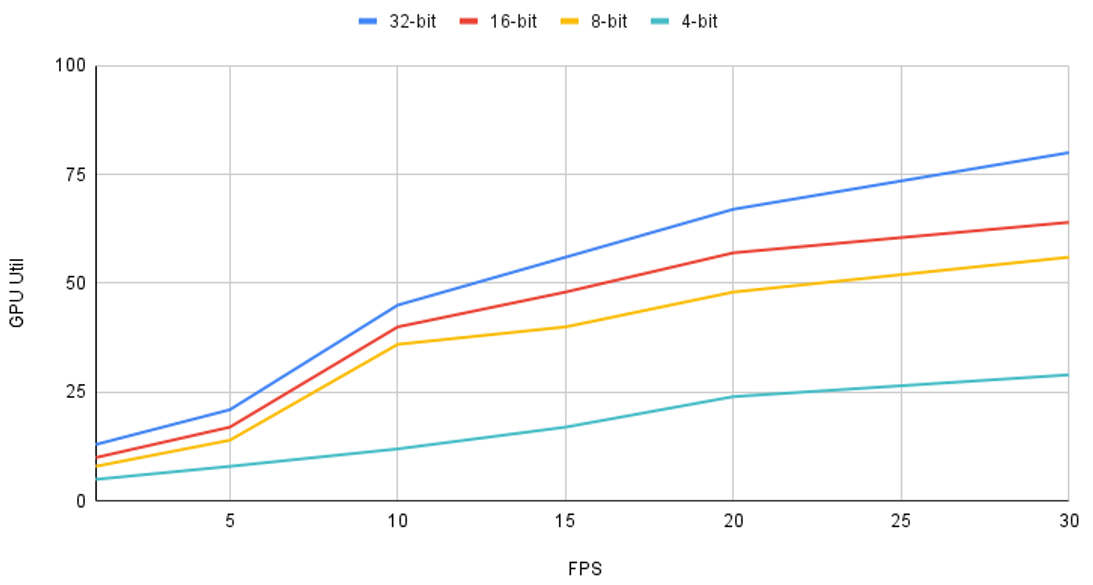}
   \caption{GPU utilization across quantization levels at varying frame rates.}
   \label{fig:quant}
\end{figure}

We deployed our best generalized model, CNN+LSTM+Intent, to iPhone 13 using CoreML \cite{coreml}, as shown in Fig. \ref{fig:deploy}. Primary objectives were to minimize inference latency and on-device resource usage, including memory, GPU, and battery. We tuned the video frame rate and conducted quantization experiments, monitoring the device’s resource consumption metrics. Fig. \ref{fig:quant} summarizes the results. The final configuration used 16-bit model quantization and a 2 FPS video frame rate, effectively balancing performance and resource efficiency. 

At any given moment, the user may choose to override the navigation command and proceed in a different direction. The model generates a prediction for every frame, resulting in a 2Hz prediction frequency. This allows the system to assess the latest state of the environment based on the user’s most recent movement, approximately every 0.5 seconds. Given the walking speeds and environments of our target users, we determined that a 2Hz prediction rate is sufficient for reliable real-time navigation while maintaining accuracy, adaptability and efficiency.
\newline

\noindent \textbf{User Privacy}: In the deployed user app, all processing occurs locally on the user’s smartphone to ensure privacy. The app transmits only anonymized model performance metrics to the server, without storing or sharing raw camera or sensor data. During data collection, video recording is conducted only in public spaces where prior consent has been obtained from the relevant authorities, ensuring compliance with ethical and legal requirements.

\section{Discussion}

In this work, we present AIGD to empower blind individuals to navigate diverse environments with greater autonomy, reducing reliance on caregivers and fostering independence. Our smartphone based deployment democratizes access to navigation assistance by reducing financial and technical barriers to adoption. Our goal is to facilitate comprehensive daily life integration by being able to support navigation across both indoor and outdoor environments.

To evaluate robustness, we assessed the model across diverse settings, demonstrating strong generalization from training environments (e.g., libraries, university halls) to unseen test locations like grocery stores. Future work will expand the dataset to more indoor and outdoor scenarios to further test performance in unfamiliar environments. Addressing this is critical to ensure user safety, especially in ambiguous situations.  

Our novel intent-conditioning technique to extend egocentric path prediction models utilize simple architectures, commonly used in literature, guided by the system's latency constraints. Future research could explore the potential of this technique with more advanced architectures and improved latency optimization.

The exploration of larger, more complex models could also be facilitated by expanding the dataset to include additional scenarios. Incorporating videos from diverse smartphone cameras would also improve the model’s invariance and generalization across different hardware configurations.

\subsection*{Error Analysis and Future Improvements}

An analysis of the most common error patterns in the model’s predictions revealed challenges in: 1) Dynamic environments, such as mis-predictions around moving objects (e.g., pedestrians) or stopping for obstacles and traffic lights, and 2) Ambiguous path structures, including blocked or forked paths and nuanced turns that are not strictly LEFT or RIGHT.

Currently, our model supports only three directional classes, but futire work could easily introduce more granular classes, such as finer-grained turning angles and start/stop walking instructions, by collecting and labeling more relevant data.

Our research into cane-walking patterns and social dynamics of blind navigation guided our focus on predicting paths based on the limited scene information captured by a smartphone camera. This relies on the assumption that canes help detect obstacles and navigate blocked paths. However, explicitly modeling the behavior of environmental entities—such as pedestrians, vehicles, and other obstacles—could enable more nuanced navigation paths. Addressing these in future work will bridge the gap between current capabilities and real-world needs.

\subsection*{Implications of the Egocentric Videos Dataset}
Our released indoor and outdoor egocentric video dataset not only advances research in first-person view analysis, ego-motion estimation, and trajectory forecasting but also has broader applications in mobility patterns, pedestrian behavior, and accessibility design.

The dataset provides valuable insights into how visually impaired individuals navigate urban environments, offering potential contributions to urban planning and infrastructure development such as: 1) Identification of challenging navigation areas that need infrastructure improvements, 2) Optimizing pedestrian pathways (e.g., sidewalks and crosswalks) for better accessibility, 3) Enhancing public transit integration by providing real-time accessibility information, such as guiding users to transit stops or optimizing wayfinding in large transport hubs.

Collaboration with public policy researchers, healthcare professionals, and city planners can further expand the dataset’s impact by: 1) Developing safety guidelines for shared spaces, 2) Informing emergency evacuation protocols for visually impaired individuals, 3) Guiding the design of mobility training programs and therapeutic interventions to enhance independent navigation.

By leveraging these insights, we can drive AI-powered accessibility innovations that improve urban mobility for visually impaired individuals and contribute to more inclusive, human-compatible AI systems.

\section{Conclusion}
This paper introduces AI Guide Dog (AIGD), an egocentric navigation system for visually impaired individuals. By introducing a novel intent-conditioning technique and a multi-label classification framework, we address key challenges such as goal-based navigation in outdoor settings and directional uncertainty in exploratory scenarios. Our approach balances simplicity and performance, enabling generalization across diverse environments while remaining practical for real-time deployment on smartphones. Additionally, our released egocentric video dataset provides valuable insights that could help further research in assistive AI and accessibility-focused urban planning. Given the limited research in this domain, we hope this work inspires further advancements in assistive navigation technologies.

\section{Acknowledgments}
We thank Anirudh Koul for his guidance and mentorship, as well as the AI Guide Dog mentorship team—Ihor Markevych, Siddha Ganju, and Toufik Zitouni—for their expertise and support. Additionally, we acknowledge the MCDS program and the Language Technologies Institute at Carnegie Mellon University for providing the resources and support that made this work possible.

\small
\bibliography{main}

\end{document}